\documentclass[twocolumn,envcountsect]{svjour3}          
\smartqed  
\usepackage{graphicx}
\usepackage{mathptmx}      
\usepackage{hyperref}
\usepackage{amsfonts}
\usepackage{amsmath}
\usepackage{algorithm}
\usepackage{algorithmic}

\journalname{Cogn Comput}
\begin{document}

\title{Anatomical Pattern Analysis for decoding visual stimuli in human brains}



\author{Muhammad Yousefnezhad \and Daoqiang Zhang}


\institute{The authors are with the College of Computer Science and Technology, Nanjing University of Aeronautics and Astronautics, Nanjing 211106, China.\\	
 \email{myousefnezhad@nuaa.edu.cn; dqzhang@nuaa.edu.cn}  
}

\date{Received: 31 Mar 2017 / Accepted: 04 Oct 2017}

\maketitle

\begin{abstract}
\textit{Background:} A universal unanswered question in neuroscience and machine learning is whether computers can decode the patterns of the human brain. Multi-Voxels Pattern Analysis (MVPA) is a critical tool for addressing this question. However, there are two challenges in the previous MVPA methods, which include decreasing sparsity and noise in the extracted features and increasing the performance of prediction. 

\textit{Methods:} In overcoming mentioned challenges, this paper proposes Anatomical Pattern Analysis (APA) for decoding visual stimuli in the human brain. This framework develops a novel anatomical feature extraction method and a new imbalance AdaBoost algorithm for binary classification. Further, it utilizes an Error-Correcting Output Codes (ECOC) method for multiclass prediction. APA can automatically detect active regions for each category of the visual stimuli. Moreover, it enables us to combine homogeneous datasets for applying advanced classification. 

\textit{Results and Conclusions:} Experimental studies on 4 visual categories (words, consonants, objects and scrambled photos) demonstrate that the proposed approach achieves superior performance to state-of-the-art methods. 

\end{abstract}

\section{Introduction}
In order to decode visual stimuli in the human brain, Multi-Voxel Pattern Analysis (MVPA) technique \cite{1norman06,40tony16,41tony17a} must apply machine learning methods to task-based functional Magnetic Resonance Imaging (fMRI) datasets. Indeed, analyzing the patterns of visual objects is one of the most interesting topics in MVPA, which can enable us to understand how brain stores and processes the visual stimuli \cite{2haxby14,3osher15}. Technically, there are two challenges in the previous studies. As the first issue, trained features are sparse and noisy because most of the previous studies in whole-brain analysis directly utilized raw voxels for predicting the stimuli \cite{2haxby14,3osher15,4friston03}. As the second challenge, improving the performance of prediction is so hard because task-based fMRI datasets can be considered as the imbalanced classification problems. For instance, consider collected data with 10 same size categories. Since this dataset is imbalance for (one-versus-all) binary classification, most of the classical algorithms cannot provide acceptable performance \cite{2haxby14,5cox03,9liu09}. 

As the main contributions, this paper proposes Anatomical Pattern Analysis (APA) for decoding visual stimuli in the human brain. To generate a normalized view, APA automatically detects the active regions and then extracts features based on the brain anatomical structures. Indeed, the normalized view can enable us to combine homogeneous datasets, and it can decrease noise, sparsity, and error of learning. Further, this paper develops a modified version of imbalance Adapting Boosting (AdaBoost) algorithm for binary classification. This algorithm uses a supervised random sampling and penalty values, which are calculated by the correlation between different classes, for improving the performance of prediction. This binary classification will be used in a one-versus-all ECOC method as a multiclass approach for classifying the categories of the brain response. 

The rest of this paper is organized as follows: The related works are presented in Section 2. This paper introduces the proposed method in Section 3. Experimental results are reported in Section 4; and finally, this paper presents conclusion and pointed out some future works in Section 5.

\section{Related Works}
There are three different types of studies for decoding stimuli in the human brain. Pioneer studies just focused on recognizing special regions of the human brain, such as inanimate objects \cite{10malach95}, faces \cite{11kanwisher97}, visually illustration of words \cite{12cohen00}, body parts \cite{13liesegang02}, and visual objects \cite{14haxby01}. Although they proved that different stimuli can provide distinctive responses in the brain regions, they cannot find the deterministic locations (or patterns) related to each category of stimuli.

The next group of studies developed correlation techniques in order to understand the similarity (or difference) between distinctive stimuli. Haxby et al. employed brain patterns located in Fusiform Face Area (FFA) and Parahippocampal Place Area (PPA) in order to analyze correlations between different categories of visual stimuli, i.e. gray-scale images of faces, houses, cats, bottles, scissors, shoes, chairs, and scrambled (nonsense) photos \cite{14haxby01}. Kamitani and Tong studied the correlations of low-level visual features in the visual cortex (V1–V4) \cite{15kamitani05}. In similar studies, Haynes et al. analyzed distinctive mental states \cite{16haynes06} and more abstract brain patterns such as intentions \cite{17haynes07}. Kriegeskorte et al. proposed Representational Similarity Analysis (RSA) in order to evaluate the similarities (or differences) among distinctive brain states \cite{18kriegeskorte08}. Connolly et al. utilized RSA in order to compare the correlations between human brains and monkey brains \cite{19connolly12,20connolly12}. RSA demonstrates that the representations of each category of stimuli in distinctive brain regions have a different structure \cite{18kriegeskorte08,19connolly12,20connolly12}. Rice et al. proved that not only the brain responses are different based on the categories of the stimuli but also they are correlated based on different properties of the stimuli. They extracted the properties of visual stimuli (photos of objects) and calculated the correlations between these properties and the brain responses. They separately reported the correlation matrices for different human faces and different objects (houses, chairs, etc.) \cite{21rice14}. 

The last group of studies proposed the MVPA techniques for predicting the category of visual stimuli. Cox et al. utilized linear and non-linear versions of Support Vector Machine (SVM) algorithm \cite{5cox03}. In order to decode the brain patterns, some studies \cite{22carlson03,23otoole05,39xu13} employed classical feature selection (ranking) techniques, such as Principal Component Analysis (PCA) \cite{22carlson03}, Linear Discriminant Analysis (LDA) \cite{23otoole05}, or Independent Component Analysis (ICA) \cite{39xu13}, that these method are mostly used for analyzing rest-state fMRI datasets. Recent studies proved that not only these techniques cannot provide stable performance in the task-based fMRI datasets \cite{24chen15,25chen16} but also they had spatial locality issue, especially when they were used for whole brain functional analysis \cite{25chen16}. Norman et al. argued for using SVM and Gaussian Naive Bayes classifiers \cite{1norman06}. Kay et al. studied decoded orientation, position and object category from the brain activities in visual cortex \cite{26kay08}. Mitchell et al. introduced a new method in order to predict the brain activities associated with the meanings of nouns \cite{27mitchell08}. Miyawaki et al. utilized a combination of multiscale local image decoders in order to reconstruct the visual images from the brain activities \cite{28miyawaki08}. In order to generalize the testing procedure for task-based fMRI datasets, Kriegeskorte et al. proved that the data in testing must have no role in the procedure of generating an MVPA model \cite{29kriegeskorte09}.

There are also some studies that focused on sparse learning techniques. Yamashita et al. developed Sparse Logistic Regression (SLR) in order to improve the performance of classification models \cite{30yamashita08}. Carroll et al. employed the Elastic Net for prediction and interpretation of distributed neural activity with sparse models \cite{31carroll09}. Varoquaux et al. proposed a small-sample brain mapping by using sparse recovery on spatially correlated designs with randomization and clustering. Their method is applied on small sets of brain patterns for distinguishing different categories based on a one-versus-one strategy \cite{32varoquaux12}. 

As the first modern approaches for decoding visual stimuli, Anderson and Oates applied non-linear Artificial Neural Network (ANN) on brain responses \cite{38anderson10}. McMenamin et al. studied subsystems underlie Abstract-Category (AC) recognition and priming of objects (e.g., cat, piano) and Specific-Exemplar (SE) recognition and priming of objects (e.g., a calico cat, a different calico cat, a grand piano, etc.). Technically, they applied SVM on manually selected ROIs in the human brain for generating the visual stimuli predictors \cite{6mcmenamin15}. Mohr et al. compared four different classification methods, i.e. L1/L2 regularized SVM, the Elastic Net, and the Graph Net, for predicting different responses in the human brain. They show that L1-regularization can improve classification performance while simultaneously providing highly specific and interpretable discriminative activation patterns \cite{7mohr15}. Osher et al. proposed a network (graph) based approach by using anatomical regions of the human brain for representing and classifying the different visual stimuli responses (faces, objects, bodies, scenes) \cite{3osher15}.
\begin{figure*}[h]
\centering
\includegraphics[width=0.99\textwidth]{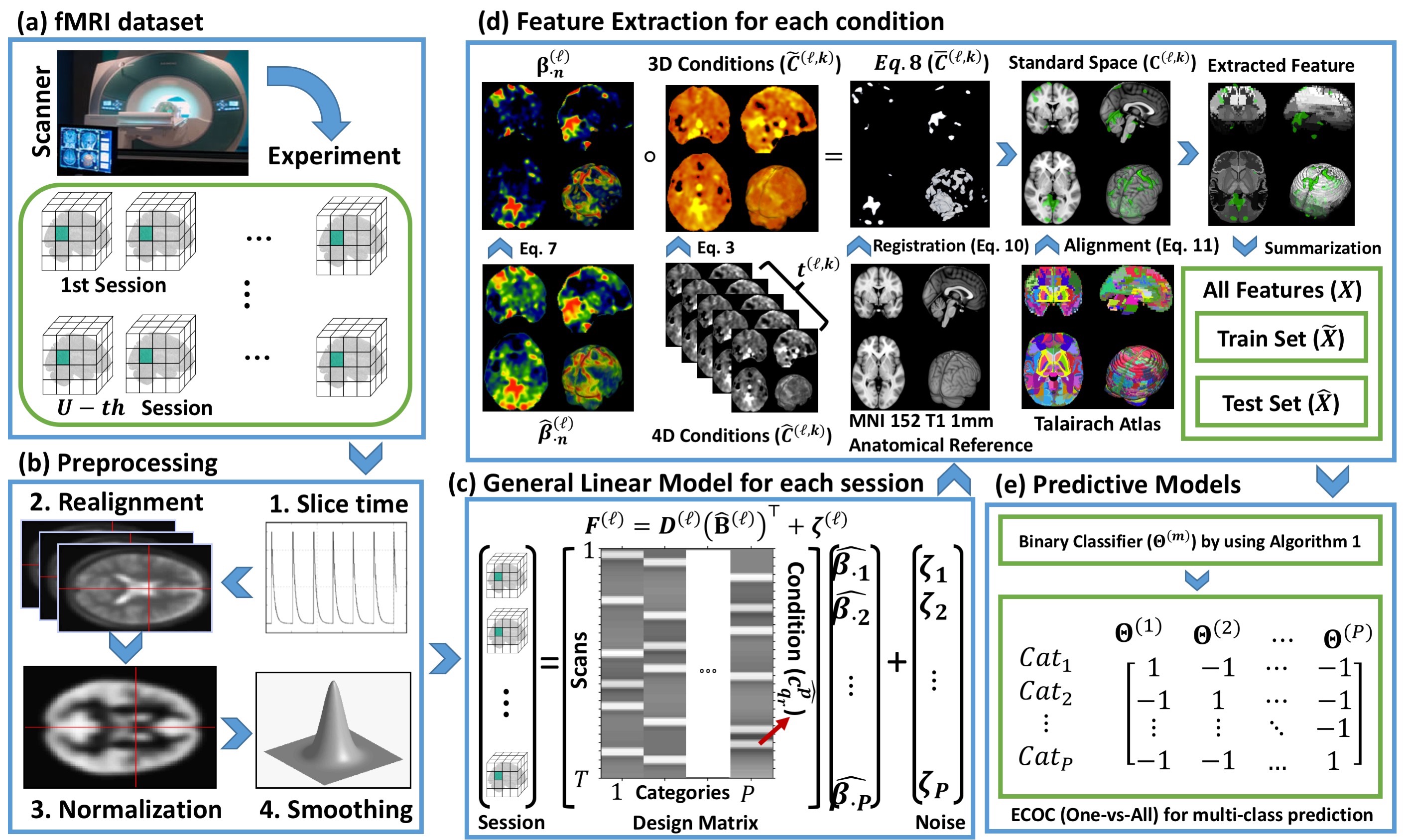}\\	
\caption{Anatomical Pattern Analysis (APA) framework}
\vskip -0.15in
\label{fig:APA}
\end{figure*}

\section{The Proposed Method}
Blood Oxygen Level Dependent (BOLD) signals are used in fMRI techniques for representing the neural activates. Based on hyperalignment problem in the brain decoding \cite{2haxby14,33haxby11,41tony17a}, quantity values of the BOLD signals in the same experiment for the two subjects are usually different. Therefore, MVPA techniques use the correlation between different voxels as the pattern of the brain response \cite{3osher15,4friston03}. As depicted in Figure \ref{fig:APA}, each fMRI experiment includes a set of sessions (time series of $3D$ images), which can be captured by different subjects or just repeating the imaging procedure with a unique subject. Technically, each session can be partitioned into a set of visual stimuli categories. Indeed, an independent category denotes a set of homogeneous conditions, which are generated by using the same type of photos as the visual stimuli. For instance, if a subject watches 6 photos of cats and 5 photos of houses during a unique session, this $4D$ image includes 2 different categories and 11 conditions.

\subsection{Feature Extraction}
Preprocessed fMRI time series collected for $U$ sessions can be defined by $\mathbf{F}^{(\ell)}=\big\{f_{mn}\big\}\in\mathbb{R}^{T\times \widehat{V}}\text{, }\ell=1\text{:}U\text{, } m=1\text{:}T\text{, }n=1\text{:}\widehat{V}$, where $T$ is the number of time points, $\widehat{V}$ denotes the number of voxels in the original space, and $f_{mn}\in\mathbb{R}$ defines the functional activity for the $\ell\text{-}th$ session in $m\text{-}th$ time point and $n\text{-}th$ voxel. Indeed, $3D$ images (tensors) in fMRI time series are considered vectorized for simplicity \cite{15kamitani05}. In addition, onsets (or time series) in the $\ell\text{-}th$ session is defined as follows:
\begin{equation}\label{eq:Onset}
\begin{split}
\mathbf{S}^{(\ell)}\in\mathbb{R}^{T\times P}=\big\{\mathbf{S}^{(\ell,1)}, \mathbf{S}^{(\ell,2)}, \dots, \mathbf{S}^{(\ell,k)}, \dots, \mathbf{S}^{(\ell,Q)}\big\}
\end{split}
\end{equation}
Here, $P$ denotes the number of categories of visual stimulus, $Q\geq P$ is the number of stimuli, the vector $\mathbf{S}^{(\ell,k)}\in\mathbb{R}^{t^{(\ell,k)}}$ denotes the onsets belonged to $k\text{-}th$ condition, and $t^{(\ell,k)}$ is the number of time points for this condition, where $t^{(\ell,k)}\ll T$. By considering (1), all time points for $k\text{-}th$ condition can be also defined as follows:
\begin{equation}
\begin{split}
\mathbf{\widehat{C}}^{(\ell,k)}=\Big\{\widehat{c}^{(\ell,k)}_{mn}\Big\}\in\mathbb{R}^{{t}^{(\ell,k)}\times\widehat{V}}=\Big\{{f}_{m.}^{(\ell)} \Big| {f}_{m.}^{(\ell)} \in \mathbf{F}^{(\ell)}, m \in\mathbf{S}^{(\ell,k)},\\ \big|\mathbf{S}^{(\ell,k)}\big| = {t}^{(\ell,k)} \Big\}.
\end{split}
\end{equation}
This paper employs maximum functional activities in each voxel as the $k\text{-}th$ condition:  
\begin{equation}
\begin{split}
\mathbf{\widetilde{C}}^{(\ell,k)}\in\mathbb{R}^{\widehat{V}}=\Big\{\max_{m=1}^{{t}^{(\ell,k)}}\big(\widehat{c}^{(\ell,k)}_{mn}\big) \Big| \widehat{c}^{(\ell,k)}_{mn} \in \mathbf{\widehat{C}}^{(\ell,k)}, n \in [1,\widehat{V}] \Big\}^\top.
\end{split}
\end{equation}
In order to extract active voxels and then automatically define Region of Interests (ROIs), $\mathbf{F}^{(\ell)}$ can be also written as a general linear model: 
\begin{equation}
\begin{split}
\mathbf{F}^{(\ell)} = \mathbf{D}^{(\ell)}\Big(\mathbf{\widehat{B}}^{(\ell)}\Big)^\top+\mathbf{\zeta}^{(\ell)},
\end{split}
\end{equation}
where $\mathbf{D}^{(\ell)}=\Big\{{d}_{mn}^{(\ell)}\Big\}\in\mathbb{R}^{T\times P}$ denotes the design matrix, $\mathbf{\zeta}^{(\ell)}$ is the noise (error of estimation), and also $\mathbf{\widehat{B}}^{(\ell)}=\Big\{{\widehat{\beta}}_{mn}^{(\ell)}\Big\}\in\mathbb{R}^{\widehat{V}\times P}$ denotes the set of correlations for $\ell\text{-}th$ session. Here, ${d}_{.n}^{(\ell)}\in\mathbb{R}^{T}$ and ${\widehat{\beta}}_{.n}^{(\ell)}\in\mathbb{R}^{\widehat{V}}$ respectively denote the design vector and voxel correlations belonged to $n\text{-}th$ category of stimuli. Now, design matrix can be calculated by: 
\begin{equation}\label{eq:DM}
\begin{split}
\mathbf{D}^{(\ell)}=\mathbf{S}^{(\ell)} * \mathcal{H}
\end{split}
\end{equation}
where $\mathcal{H}$ is the Hemodynamic Response Function (HRF) signal \cite{4friston03}, and $*$ denotes the convolution operator. In order to solve (4), this paper uses Generalized Least Squares (GLS) \cite{4friston03} approach as follows: 
\begin{equation}
\begin{split}
\mathbf{\widehat{B}}^{(\ell)}=\bigg(\Big(\big(\mathbf{D}^{(\ell)}\big)^\top\big({\Sigma}^{(\ell)}\big)^{-1}\mathbf{D}^{(\ell)}\Big)^{-1}\big(\mathbf{D}^{(\ell)}\big)^\top\big({\Sigma}^{(\ell)}\big)^{-1}\mathbf{F}^{(\ell)}\bigg)^\top,
\end{split}
\end{equation}
where ${\Sigma}^{(\ell)}$ is the covariance matrix of the noise (var$\big(\zeta^{(\ell)}\big)={\Sigma}^{(\ell)}\sigma^2\neq\mathbf{I}\sigma^2$) \cite{2haxby14,4friston03,34tony17b}. Further, activated voxels can be defined as follows by using the positive values of the correlation matrix:
\begin{equation}
\begin{split}
\mathbf{B}^{(\ell)}=\big\{\beta_{mn}^{(\ell)}\big\}\in\mathbb{R}^{V\times P}=  
\begin{cases}
\widehat{\beta}_{mn}^{(\ell)}       & \quad \widehat{\beta}_{mn}^{(\ell)} > 0\\
0  									& \quad \text{otherwise}\\
\end{cases},
\end{split}
\end{equation}
where $\widehat{\beta}_{mn}^{(\ell)}\in\mathbf{\widehat{B}}^{(\ell)}$ denotes the estimated correlation matrix in \eqref{eq:DM}. Indeed, non-zero elements in $\beta_{.n}^{(\ell)}$ are all activated voxels belonged to $n\text{-}th$ category of stimuli. These activated voxel correlations can be applied to the conditions as follows:
\begin{equation}\label{eq:ActiveVoxels}
\begin{split}
\mathbf{\bar{C}}^{(\ell,k)}\in\mathbb{R}^{\widehat{V}}=\mathbf{\widetilde{C}}^{(\ell,k)}\circ\beta_{.n}^{(\ell)}
\end{split}
\end{equation}
where $\circ$ denotes Hadamard product. Here, the $k\text{-}th$ condition must be belonged to the $n\text{-}th$ category of stimuli. Since mapping whole of fMRI time series to standard space decreases the performance of the final results, most of the previous studies use the original images instead of the standard version. By considering \eqref{eq:ActiveVoxels} for each condition, this paper enables to map brain activities to a standard space. This mapping can provide normalized view for combing homogeneous datasets. For registering \eqref{eq:ActiveVoxels} to standard space, this paper utilizes the fMRI Linear Image Registration Tool (FLIRT) algorithm \cite{35jenkinson02}, which minimizes the following cost function:
\begin{equation}
\begin{split}\label{eq:Reg}
\tau^{(\ell,k)}\in\mathbb{R}^{V\times\widehat{V}}=\arg\min\Big(\text{NMI}\big(\mathbf{\bar{C}}^{(\ell,k)} ,\mathcal{R}\big)\Big)
\end{split}
\end{equation}
where $\mathcal{R}$ denotes the reference image, the function $\text{NMI}$ is the Normalized Mutual Information between two images, $\tau^{(\ell,k)}$ denotes the transformation matrix. The performance of \eqref{eq:Reg} will be analyzed in Section 4. Further, the final mapping can be also defined as follows:
\begin{equation}
\begin{split}
\tau^{(\ell,k)}\text{:}\mathbb{R}^{\widehat{V}}\to\mathbb{R}^{{V}}\implies\mathbf{C}^{(\ell,k)}=\tau^{(\ell,k)}\mathbf{\bar{C}}^{(\ell,k)}
\end{split}
\end{equation}
where ${C}^{(\ell,k)}=\big\{{c}_{1}^{(\ell,k)}, {c}_{2}^{(\ell,k)}, \dots, {c}_{i}^{(\ell,k)}, \dots, {c}_{V}^{(\ell,k)}\big\}\text{, }{c}_{i}^{(\ell,k)}\in\mathbb{R}\text{, }V$ denotes the number of voxels in the standard space. In order to reduce the sparsity of ${C}^{(\ell,k)}$, this paper employs an anatomical atlas. Now, consider $\mathbf{A}=\big\{\mathbf{A}_1, \mathbf{A}_2,\dots,\mathbf{A}_n,\dots\mathbf{A}_E\big\}$ where $E$ is the number of atlas regions, $\bigcap_{n=1}^{E}\mathbf{A}_n=\emptyset$,\\ $\bigcup_{n=1}^{E}\mathbf{A}_n=\mathbf{A}$, and $\mathbf{A}_n$ denotes the set of indexes of voxels for the $n\text{-}th$ region. The extracted feature for $\ell\text{-}th$ session in $k\text{-}th$ condition and $n\text{-}th$ anatomical region is calculated as follows: 
\begin{equation}\label{eq:Fea}
\begin{split}
x^{(\ell,k,n)}\in\mathbb{R}=\frac{1}{\big|\mathbf{A}_n\big|}\sum_{i\in\mathbf{A}_n}{c}_{i}^{(\ell,k)}
\end{split}
\end{equation}
Finally, the extracted features for each condition can be defined by $\mathbf{X}^{(\ell,k)}\in\mathbb{R}^{E}=\big\{x^{(\ell,k,1)}, x^{(\ell,k,2)}, \dots, x^{(\ell,k,n)}, \dots, x^{(\ell,k,E)}\big\}$, where $E<V$. Here, $\mathbf{X}^{\ell}\in\mathbb{R}^{E\times Q}$ denotes the extracted features for the $\ell\text{-}th$ session. In addition, whole of dataset can be defined by $\mathbf{X}\in\mathbb{R}^{E\times QU}$, where each column denotes the extracted features for an individual stimulus.
\begin{figure}[t]
	\centering
	\includegraphics[width=0.48\textwidth]{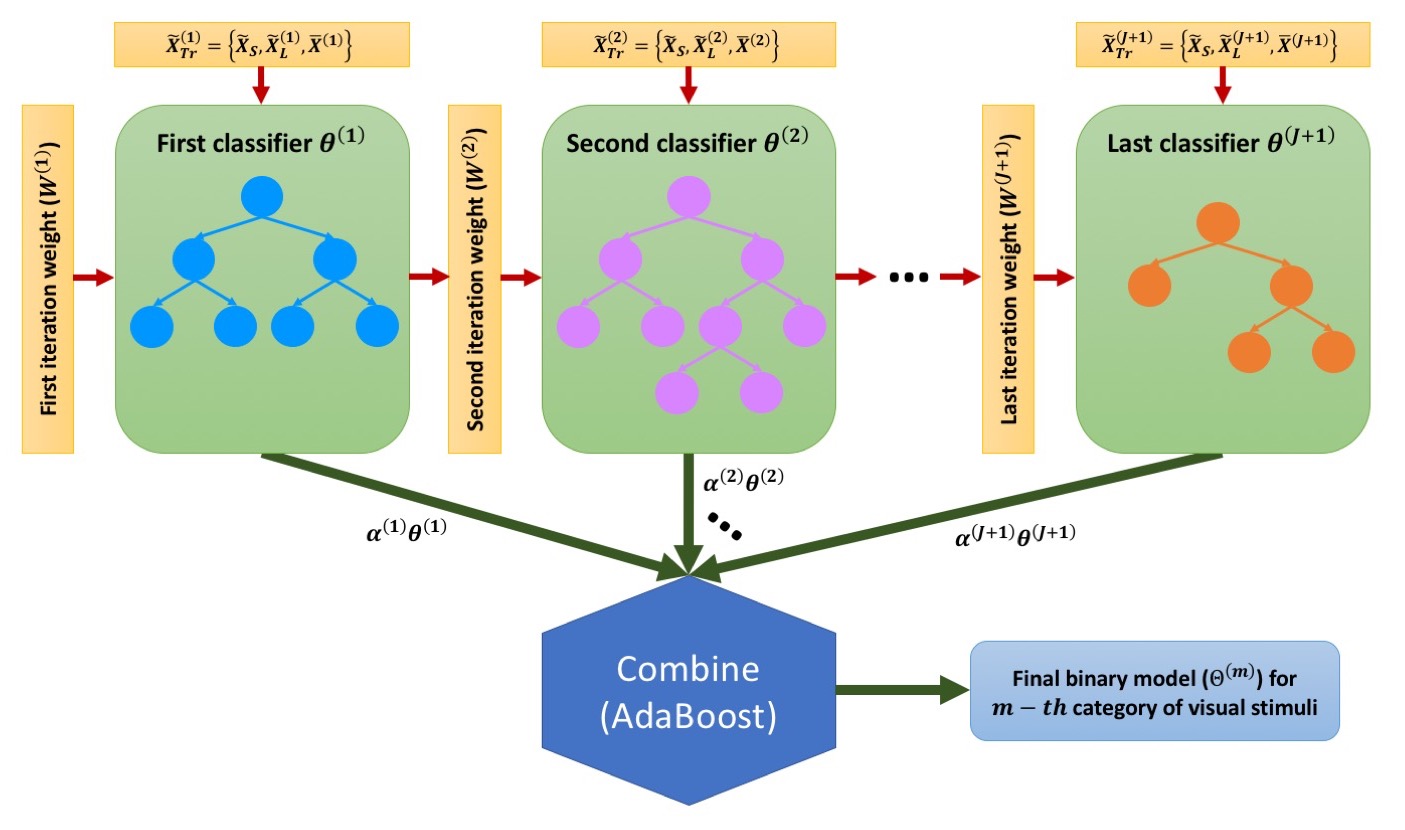}\\	
	\caption{The proposed AdaBoost algorithm for applying a robust binary classification}
	\vskip -0.15in
	\label{fig:AdaBoost}
\end{figure}
\subsection{Binary Classification Algorithm}
In previous sections, we mentioned the imbalance issue in the MVPA analysis. In practice, there are two approaches in order to deal with this issue, i.e. designing an imbalance classifier, or converting the imbalance problem to an ensemble of balance classification models. Previous studies demonstrated that the performance of imbalance classifiers may not be stable, especially when we have sparsity and noise in our datasets \cite{9liu09,34tony17b,40tony16,41tony17a}. Since fMRI datasets mostly include noise and sparsity, this paper has chosen the ensemble approach. Technically, ensemble learning also contains two groups of solutions, i.e. bagging or boosting. While bagging generates all classifiers at the same time and then combine all of them as the final model, the boosting gradually creates each classifier in order to improve the performance of each iteration by tracing errors of previous iterations. We just have to note that ensemble learning can be used in both balance and imbalance problems. In fact, the main difference comes from the strategy of sampling. In balance problems, sampling methods are applied to the whole of datasets, whereas instances of the large class are sampled in the imbalance problems \cite{9liu09}. As depicted in Figure \ref{fig:AdaBoost}, this paper presents a new branch of AdaBoost algorithm in order to significantly improve the performance of the final model in fMRI analysis. In a nutshell, this algorithm firstly converts an imbalance MVPA problem to a set of balance problems. Then, it iteratively applies the decision tree \cite{9liu09} to each of these balance problems. Finally, AdaBoost is used in order to generate the final model. In the proposed method, the weight of each classifier (tree) for the final combination is generated based on the error (failed predictions) of the previous iterations for gradually improving the performance of the final model. 

\begin{algorithm}[!t]
	\caption{The proposed binary classification algorithm}
	\label{alg:Binary}
	\begin{algorithmic}
		\STATE {\bfseries Input:} Training set $\mathbf{\widetilde{X}}$, Class labels $\mathbf{Y}^{(m)}$.\\
		\STATE {\bfseries Output:}  Set of classifiers $\Theta^{(m)}$.\\
		\STATE {\bfseries Method:}\\
		01. Based on $\mathbf{Y}^{(m)}$, partitioning $\mathbf{\widetilde{X}}=\{\mathbf{\widetilde{X}}_S, \mathbf{\widetilde{X}}_L\}$\\
		02. Calculating $J=\text{int}(\frac{|\mathbf{\widetilde{X}}_L|}{|\mathbf{\widetilde{X}}_S|})$.\\
		03. Random sampling: $\mathbf{\widetilde{X}}_L=\big\{\mathbf{\widetilde{X}}_{L}^{(1)}, \mathbf{\widetilde{X}}_{L}^{(2)}, \dots, \mathbf{\widetilde{X}}_{L}^{(n)}, \dots, \mathbf{\widetilde{X}}_{L}^{(J)}\big\}$.\\
		04. Initiate $\mathbf{\bar{X}}^{(1)}=\mathbf{\bar{Y}}^{(1)}=\emptyset$.\\
		05. \textbf{For} $(n=1 \dots J)$:\\
		06. \quad $\mathbf{\widetilde{X}}^{(n)}_{Tr}=\big\{\mathbf{\widetilde{X}}_S, \mathbf{\widetilde{X}}_{L}^{(n)}, \mathbf{\bar{X}}^{(n)}\big\}$ as training-set for this iteration.\\
		07. \quad $\mathbf{\widetilde{Y}}^{(n)}_{Tr}=\big\{\mathbf{\widetilde{Y}}_S, \mathbf{\widetilde{Y}}_{L}^{(n)}, \mathbf{\bar{Y}}^{(n)}\big\}$ as class labels for this iteration.\\
		08. \quad$\mathbf{W}^{(n)}=
		\begin{cases}
		1 &\quad \text{for instances of } \mathbf{\widetilde{X}}_S \text{ or } \mathbf{\bar{X}}^{(n)}\\
		1 - \big|\text{corr}\big(\mathbf{\widetilde{X}}_S, \mathbf{\widetilde{X}}_{L}^{(n)}\big)\big| &\quad \text{for instances of } \mathbf{\widetilde{X}}_{L}^{(n)}\\
		\end{cases}$\\
		09. \quad  $\theta^{(n)}=\text{classifier}\big(\mathbf{\widetilde{X}}^{(n)}_{Tr},\mathbf{\widetilde{Y}}^{(n)}_{Tr}, \mathbf{W}^{(n)}\big)$ as weighted decision tree.\\
		10 \quad Constructing $\mathbf{\bar{X}}^{(n+1)}$ as instances cannot truly trained in $\theta^{(n)}$.\\
		11. \quad $\epsilon^{(n)}=\frac{|\mathbf{\bar{X}}^{(n+1)}|}{|\mathbf{\widetilde{X}}^{(n)}_{Tr}|}$ as error of classification.\\
		12. \quad $\alpha^{(n)}=\frac{1}{2}\ln\big(\frac{1-\epsilon^{(n)}}{\epsilon^{(n)}}\big)$ AdaBoost weight for the classifier $\theta^{(n)}$.\\
		13. \textbf{End For}\\
		14. \textbf{Return} $\Theta^{(m)}\big(x\big)=\text{sign}\Big(\sum_{n=1}^{J+1}\alpha^{(n)}\theta^{(n)}\big(x\big)\Big)$ as the final model.
	\end{algorithmic}
\end{algorithm}

In order to apply the binary classification, this paper randomly partitions the extracted features $\mathbf{X}$ into the training set $\mathbf{\widetilde{X}}$ and the testing set $\mathbf{\widehat{X}}$. As a new branch of AdaBoost algorithm, Algorithm \ref{alg:Binary} employs $\mathbf{\widetilde{X}}$ for training binary classification. Then, $\mathbf{\widehat{X}}$ is utilized for estimating the performance of the final model. As mentioned before, the binary classification for fMRI analysis is mostly imbalance, especially by using a one-versus-all strategy. Consequently, the number of samples in one of these binary classes is smaller than the other classes. As previously mentioned, this paper exploits this concept in order to solve the imbalance issue. Indeed, Algorithm 1 firstly partitions the training data $\mathbf{\widetilde{X}}$ into small $\mathbf{\widetilde{X}}_S$ and large $\mathbf{\widetilde{X}}_L$ classes (groups) based on the class labels $\mathbf{Y}^{(m)}\in\big\{+1,-1\big\}$. Here, all labels are $-1$ except the label of instances belong to $m\text{-}th$ category of visual stimuli. Then, it calculates the scale $J$ of existed elements between two classes. We have to note that $\text{int}()$ defines the floor function. As the next step, the large class is randomly partitioned into $J$ parts. Indeed, $J$ is the number of balance subsets generated from the imbalance dataset. Consequently, the number of the ensemble iteration is $J$. In each balance subset, training data $\mathbf{\widetilde{X}}_{Tr}^{(n)}$ is generated by all instances of the small class $\mathbf{\widetilde{X}}_S$, one of the partitioned parts of the large class $\mathbf{\widetilde{X}}_L$, and the instances of the previous iteration $\mathbf{\bar{X}}^{(n)}$, which cannot truly be trained (the failed predictions). After that, training weights for the final combination ($\mathbf{W}^{n}\in[0,1]$) are calculated by using the Pearson correlation (corr$(a,b)=\frac{\text{cov}(a,b)}{\sigma_a\sigma_b}$) between training instances, where larger values increase the learning sensitivity. Indeed, these weights are always maximized for the instances of the small class and the failed instances of the previous iterations. Further, the weights of the other instances are a scale of the correlation between the large class and the small class. Therefore, these weights are updated in each iteration based on the performance of previous iterations. As the last step of each iteration, the proposed method generates a classification model ($\theta^{(n)}$) and its weight ($\alpha^{(n)}$) for the final combination. While $\text{classifier}()$ can denote any kind of weighted classification algorithm, this paper employs a simple weighted decision tree \cite{9liu09} as the classification model. At the end, the final model is created by applying the AdaBoost method to the generated balance classifiers.
\subsection{Multiclass Classification Algorithm}
In this paper, a multiclass classifier is a prediction model in order to map extracted features to the category of visual stimuli, i.e. $\Theta:\mathbf{\widehat{X}}\to\mathbf{Y}_{pred}$ where $\mathbf{Y}_{pred}\in\big\{1,2,\dots,P\big\}$. Generally, there are two techniques for applying multiclass classification. The first approach directly creates the classification model such as multiclass support vector machine \cite{5cox03} or neural network \cite{1norman06}. In contrast, decomposition design (indirect) uses an array of binary classifiers for solving the multiclass problems.

\begin{figure}[t]
	\centering
	\includegraphics[width=0.48\textwidth]{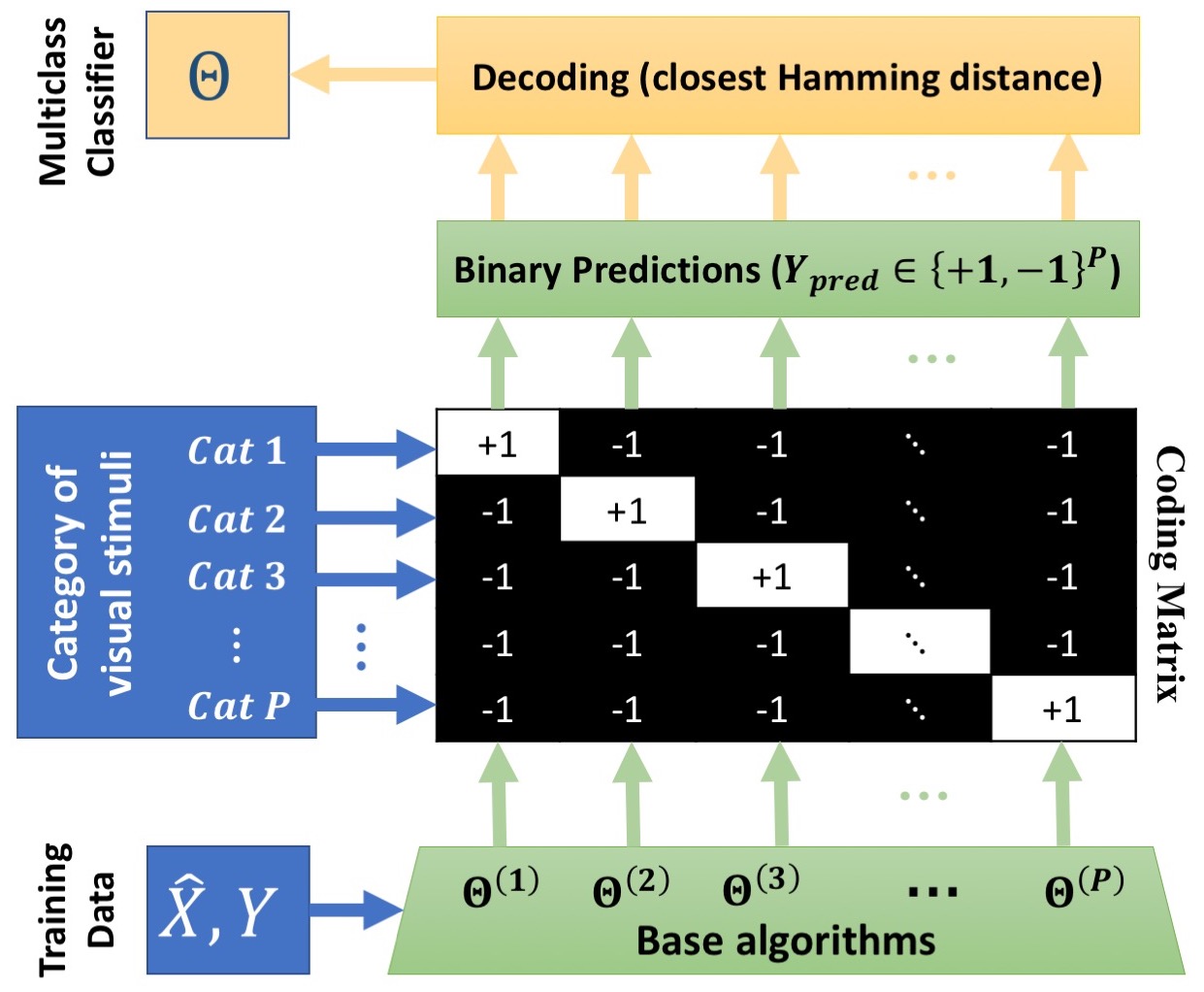}\\	
	\caption{The proposed Error-Correcting Output Codes (ECOC) approach for multiclass classification}
	\vskip -0.15in
	\label{fig:ECOC}
\end{figure}

\begin{table*}[t]
	\caption{tbl:datasets}
	\label{tbl:Datasets}
	\begin{small}
		\begin{center}	
			\begin{tabular}{lccccccccccc}
				\hline
				Title & ID & S&U&P&T&X&Y&Y&Scanner&TR&TE\\
				\hline
				Visual Object Recognition& 
				DS105& 6& 71& 8& 121& 79& 79& 75& G 3T& 2500& 30\\
				Word and Object Processing& DS107& 49& 98& 4& 164& 53& 63& 52& S 3T& 2000& 28\\
				Multi-subject, multi-modal& DS117& 20& 171& 2& 210& 64& 61& 33& S 3T& 2000& 30\\
				\hline
			\end{tabular}
		\end{center}
$S$ denotes the number of subject; $U$ is the number of sessions; $P$ denotes the number of stimulus categories; $T$ is the number of scans in unites of TRs (Time of Repetition); $X, Y, Z$ are the size of 3D images in the original space; Scanners include $S=$Siemens, and $G=$General Electric in 3 Tesla; TR is Time of Repetition in millisecond; TE denotes Echo Time in millisecond; Please see \url{openfmri.org} for more information.
	\end{small}
\end{table*}

Based on the previous discussion related to imbalance issue in fMRI datasets, this paper utilizes Error-Correcting Output Codes (ECOC) as an indirect multiclass approach in order to extend the proposed binary classifier for the multiclass prediction. As depicted in Figure \ref{fig:ECOC}, ECOC includes three components, i.e. base algorithms, coding matrix and decoding procedures \cite{8escalera10}. Since this paper uses one-versus-all encoding strategy, Algorithm \ref{alg:Binary} is employed as the based algorithms ($\Theta^{(m)}$) in the ECOC, where it generates a binary classifier for each category of visual stimuli. In other words, each independent category of the visual stimuli is compared with the rest of categories. Consequently, the size of the coding matrix is $P\times P$, where $i\text{-}th$ diagonal cell of this matrix represents the positive predictions belong to the $i\text{-}th$ category of visual stimuli and the rest of cells in this matrix determine the other categories of visual stimuli. Indeed, the number of classifiers in this strategy is exactly equal to the number of categories. As decoding stage, binary predictions, which are generated by applying the brain response to the base algorithms, are assigned to the category in the coding matrix with closest Hamming distance.

In order to present an example for ECOC procedure, consider fMRI dataset with $4$ categories of visual stimuli, i.e. photos of shoes, houses, bottles, and human faces. In this problem, $4$ different binary classifiers must be trained in order to distinguish each category of visual stimuli versus the rest of them (one-versus-all strategy). A $4\times 4$ coding matrix is also generated where each diagonal element represents the positive class of these categories (classifiers). By considering the order of the coding matrix, each prediction is assigned to the closest Hamming distance in the coding matrix. In other words, if these classifiers generate the prediction $[+1, -1, -1, -1]$ for a testing instance, then this instance definitely belongs to the first category of visual stimuli. Similarly, the prediction $[-1, +1, -1, -1]$ means the instance belongs to the second category, and etc.

\section{Results}
\subsection{Datasets}

As depicted in Table \ref{tbl:Datasets}, this paper employs $3$ datasets, shared by \url{openfmri.org}, for running empirical studies. As the first dataset, `Visual Object Recognition' (DS105) includes $P=8$ categories of visual stimuli, i.e. gray-scale images of faces, houses, cats, bottles, scissors, shoes, chairs, and scrambled (nonsense) photos. This dataset is analyzed in high-level visual stimuli as the binary predictor, by considering all categories except scrambled photos as objects, and low-level visual stimuli in the multiclass prediction. Please see \cite{2haxby14,5cox03} for more information. As the second dataset, `Word and Object Processing' (DS107) contains $P=4$ categories of visual stimuli, i.e. words, objects, scrambles, and consonants. Please see \cite{36duncan09} for more information. As the last data set, `Multi-subject, multi-modal human neuroimaging dataset' (DS117) includes MEG and fMRI images. This paper just uses the fMRI images of this dataset. It also contains $P=2$ categories of visual stimuli, i.e. human faces, and scrambles. Please see \cite{37wakeman15} for more information.

These datasets are preprocessed by SPM 12 (\url{www.fil.ion.ucl.ac.uk/spm/}), i.e. slice timing, realignment, normalization, smoothing. Further, whole-brain functional alignment is applied based on \cite{25chen16}. Then, the beta values are calculated for each session. This paper employs the \emph{MNI 152 T1 1mm} (see Figure \ref{fig:APA}.d) as the reference image ($\mathcal{R}$) in \eqref{eq:Reg} for registering the extracted conditions ($\mathbf{\bar{C}}^{(\ell,k)}$) to the standard space ($\mathbf{C}^{(\ell,k)}$). In addition, this paper uses \emph{Talairach Atlas} (contains regions) in \eqref{eq:Fea} for extracting features.

\begin{figure*}[t]
	\centering
\begin{minipage}{0.24\linewidth}
	\includegraphics[width=0.99\textwidth,height=0.9\linewidth]{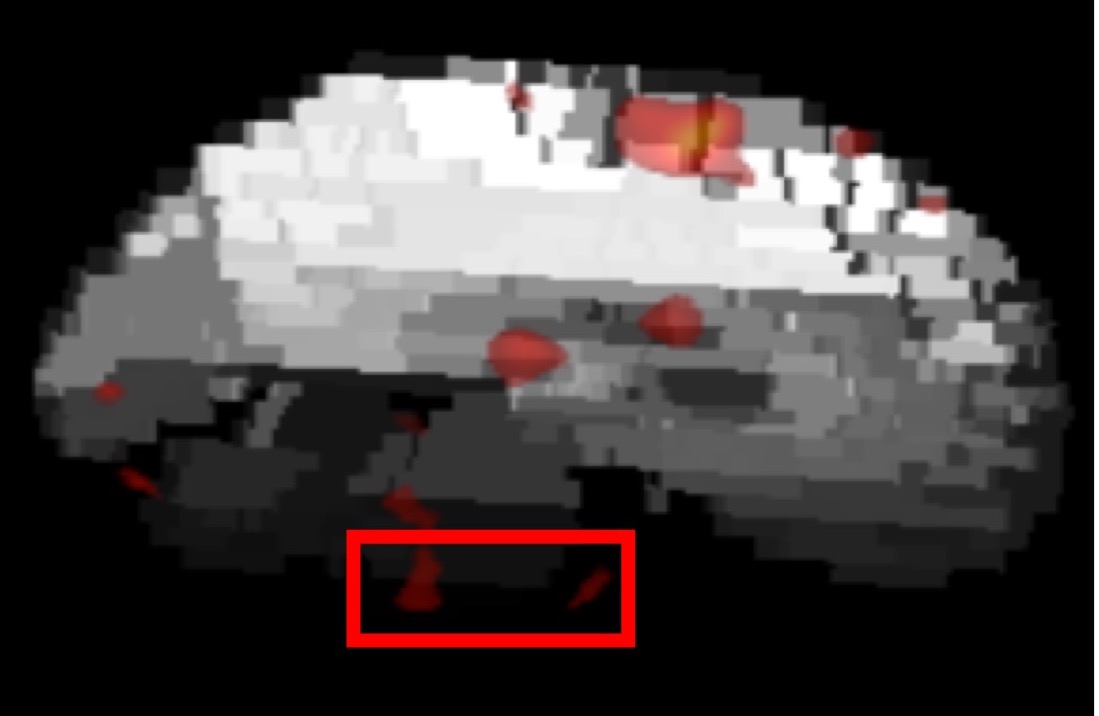}\\	
	\centering (A)
\end{minipage}
\begin{minipage}{0.24\linewidth}
	\includegraphics[width=0.99\textwidth,height=0.9\linewidth]{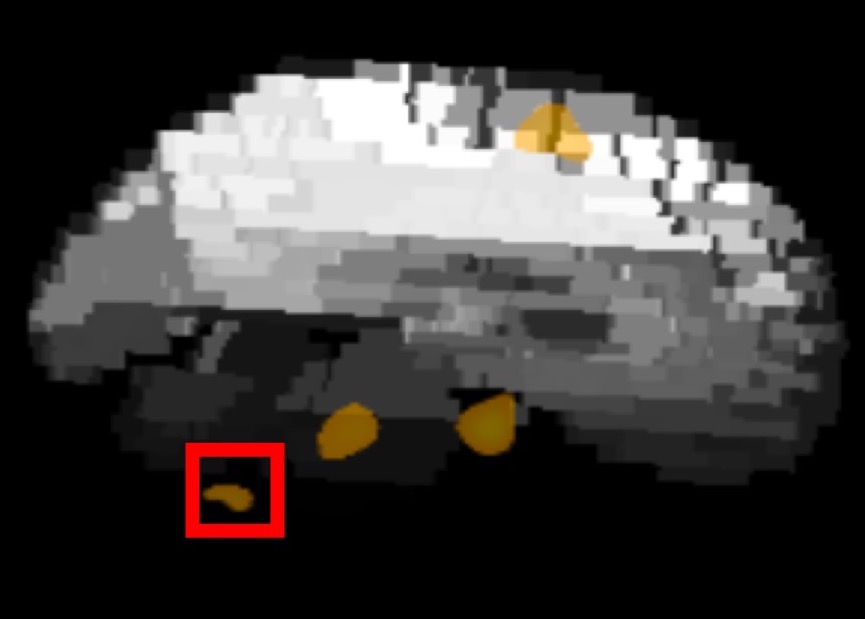}\\	
	\centering (B)
\end{minipage}
\begin{minipage}{0.24\linewidth}
	\includegraphics[width=0.99\textwidth,height=0.9\linewidth]{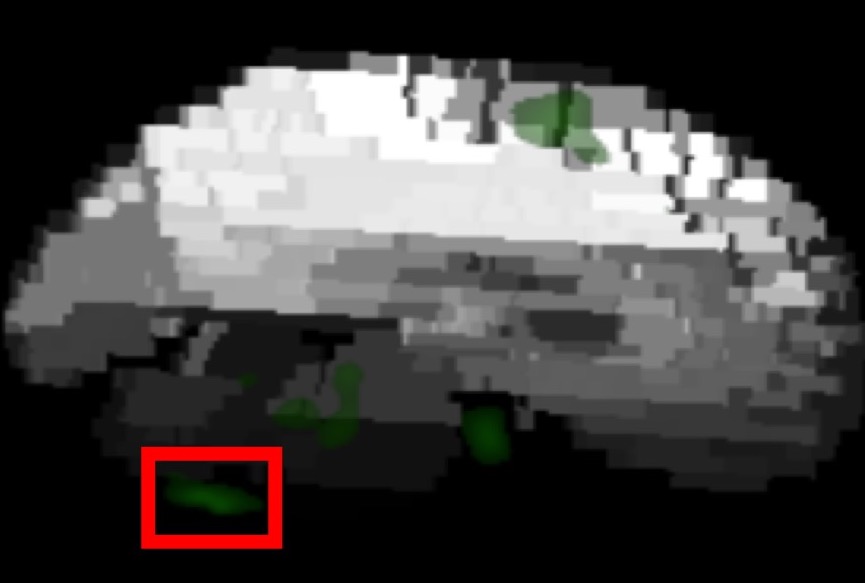}\\	
	\centering (C)
\end{minipage}
\begin{minipage}{0.24\linewidth}
	\includegraphics[width=0.99\textwidth,height=0.9\linewidth]{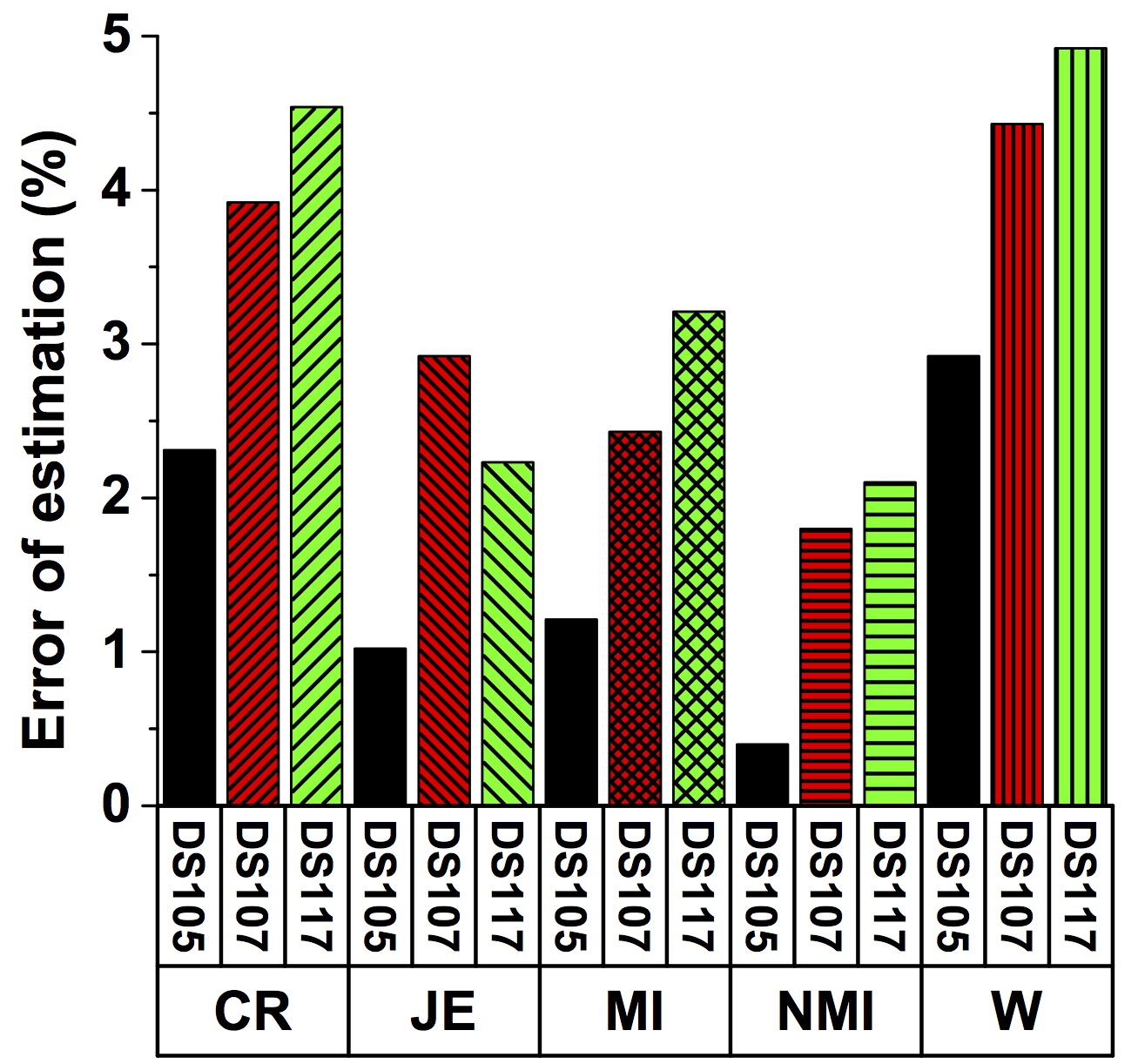}\\	
	\centering (D)
\end{minipage}
\caption{Extracted features based on different stimuli, i.e. (A) word, (B) object, and (C) scramble. (D) The effect of different objective functions in (4) on the error of registration.}
\vskip -0.15in
\label{fig:Reg}
\end{figure*}

\subsection{Parameter Analysis}
The registration objective function in \eqref{eq:Reg} will be analyzed in this section by using different distance metrics, i.e. Woods function (W), Correlation Ratio (CR), Joint Entropy (JE), Mutual Information (MI), and Normalized Mutual Information (NMI) \cite{34tony17b,35jenkinson02}. Figures \ref{fig:Reg}.a-c demonstrate examples of brain responses to different stimuli, i.e. (a) word, (b) object, and (c) scramble. Here, gray parts show the anatomical atlas, the colored parts (red, yellow and green) define the functional activities, and also the red rectangles illustrate the error areas after registration. Indeed, these errors can be formulated as the nonzero areas in the brain image which are located in the zero area of the anatomical atlas (the area without region number). Indeed, the registration errors are mostly related to the distance metrics. Previous studies illustrated that the entropy-based metrics can provide better performance \cite{34tony17b,35jenkinson02}. The performances of objective function \eqref{eq:Reg} on DS105, DS107, and DS117 datasets are analyzed in Figure \ref{fig:Reg}.d by using the mentioned distance metrics. As depicted in this figure, the entropy-based metrics (JE, MI, NMI) have provided better performance in comparison with other metrics. Since NMI uses normalization for removing the scaling effect \cite{34tony17b}, it generated the best results among the entropy-based metrics. Therefore, this paper employs NMI as the distance metric in \ref{eq:Reg} for mapping the brain activities from the original space to the standard space.

\begin{figure}[t]
	\centering
	\begin{minipage}{0.99\linewidth}
		\includegraphics[width=0.99\textwidth,height=0.7\linewidth]{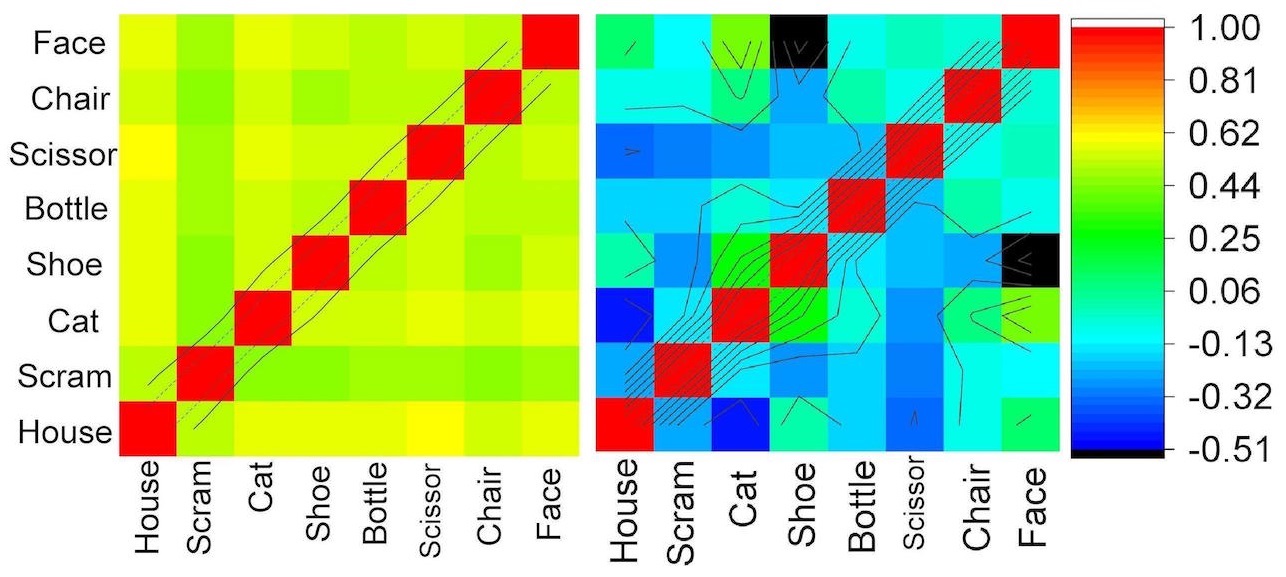}\\	
		\centering (A) \qquad\qquad\qquad (B) \\ DS105\\
	\end{minipage}
	\begin{minipage}{0.49\linewidth}
		\includegraphics[width=0.99\textwidth,height=0.9\linewidth]{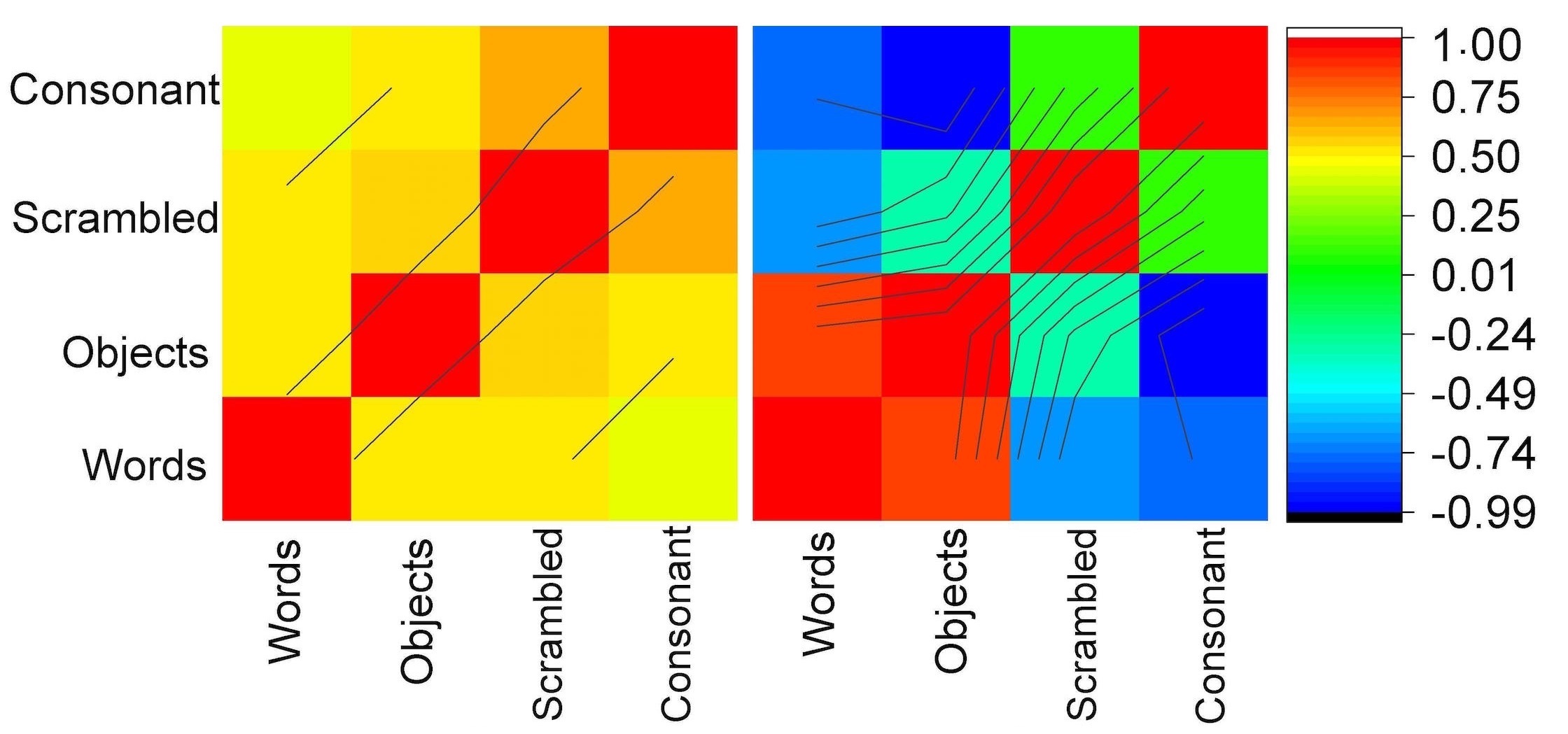}\\	
		\centering (C) \qquad (D) \\  DS107\\
	\end{minipage}
	\begin{minipage}{0.49\linewidth}
		\includegraphics[width=0.99\textwidth,height=0.9\linewidth]{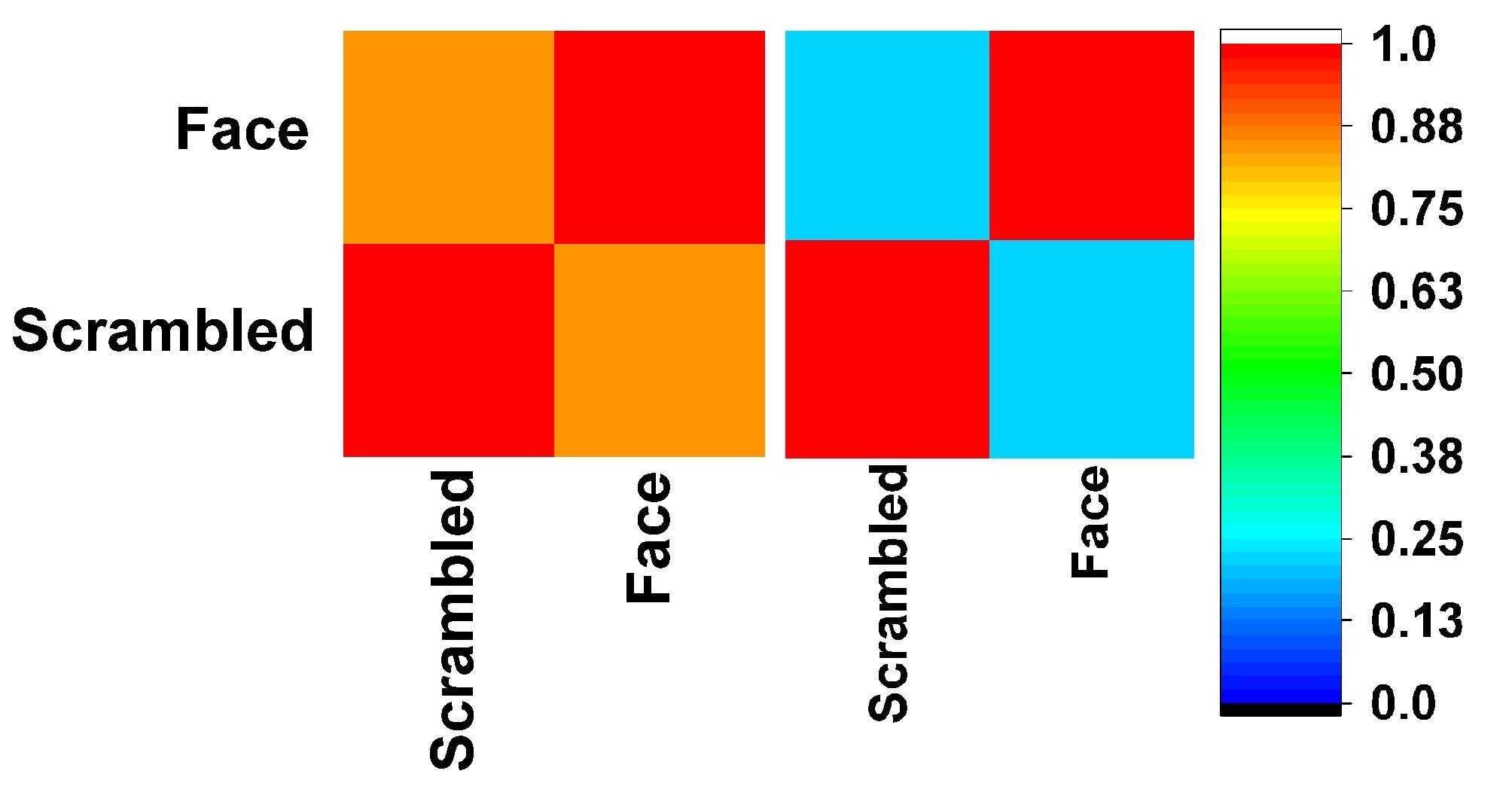}\\	
		\centering (E) \qquad (F) \\ DS117\\
\end{minipage}
	\caption{The correlation matrices: (A) raw voxels and (B) extracted features in the DS105 dataset, (C) raw voxels and (D) extracted features in the DS107 dataset, (E) raw voxels and (F) extracted features in the DS117 dataset.}
	\vskip -0.15in
	\label{fig:Corr}
\end{figure}

\subsection{Correlation Analysis}
The correlations of the extracted features will be compared with the correlations of the original voxels in this section. Previous studies illustrated that patterns of different Abstract-Categories (ACs), which is extracted from a suitable feature representation, must provide distinctive correlation values \cite{6mcmenamin15,21rice14}. Therefore, the main assumption in this section is that better feature representation (extraction) can improve the correlation analysis, where the correlation between different categories of visual stimuli must be significantly \\smaller than the correlation between stimuli belonged to the same category. In order to provide a better perspective, the extracted features are compared by considering two different levels. At the first level, the feature space is compared with the whole of raw voxels in the original space, where this comparison analyzes the correlation between whole-brain data and automatically detected ROIs. At the second level, the correlation values among different ACs are compared in the feature space that it shows how much the feature space is well-designed. 

Figure \ref{fig:Corr}.A, B, and E illustrate the correlation matrix of the DS105, DS107, and DS117 at the raw voxel space, respectively. Furthermore, Figure \ref{fig:Corr}.B, D, and F respectively show the correlation matrix the DS105, DS107, and DS117 in the feature space. As these figures depicted, different ACs are highly correlated in the voxel space. Indeed, the average of correlations is around $+0.5$ in DS105 and DS107. And, this average is around $+0.8$ in DS117 because this dataset just includes $2$ classes (photos of scramble and human face). The main reason for these results is that brain responses in the voxel space are sparse, high-dimensional and noisy. Therefore, it is so hard to discriminate between different categories (ACs) in the original space, especially when whole-brain data will be analyzed. By contrast, the feature space (Figure \ref{fig:Corr}.B, D, and E) provides distinctive representation when the proposed method used the correlated patterns in each anatomical region as the extracted features. 

\begin{table*}[!t]
\caption{Accuracy of binary predictors (mean$\pm$std)}
\label{tbl:BinaryAccuracy}
\begin{center}\begin{small}\begin{tabular}{lcccccc}
\hline
$\downarrow$Alg., Datasets$\rightarrow$ & DS105 (Objects) & DS107 (Words) & DS107 (Consonants)  & DS107 (Objects) & DS107 (Scramble) & DS117 \\
\hline
SVM \cite{5cox03} & 
71.65$\pm$0.97 & 
69.89$\pm$1.02 & 
67.84$\pm$0.82 & 
65.32$\pm$1.67 & 
67.96$\pm$0.87 & 
81.25$\pm$1.03 \\
Elastic Net \cite{7mohr15} & 
80.77$\pm$0.61 & 
78.26$\pm$0.79 & 
75.53$\pm$9.87 & 
84.15$\pm$0.89 & 
87.34$\pm$0.93 & 
86.49$\pm$0.70\\
Graph Net  \cite{7mohr15} & 
79.23$\pm$0.74 & 
79.91$\pm$0.91 & 
74.01$\pm$0.84 & 
85.96$\pm$0.76 & 
86.21$\pm$0.51 & 
85.49$\pm$0.88\\
PCA \cite{23otoole05} &\
72.15$\pm$0.76 & 
70.32$\pm$0.92 & 
69.57$\pm$1.10 & 
68.78$\pm$0.64 & 
69.41$\pm$0.35 & 
81.92$\pm$0.87\\
ICA \cite{39xu13} &
73.25$\pm$0.81 &
70.82$\pm$0.67 &
71.87$\pm$0.94 &
67.99$\pm$0.75 &
72.48$\pm$0.89 &
80.71$\pm$1.16 \\
Selected ROI \cite{6mcmenamin15}&
83.06$\pm$0.36&
89.62$\pm$0.52&
87.82$\pm$0.37&
84.22$\pm$0.44&
86.19$\pm$0.26&
85.19$\pm$0.56\\
L1 Reg. SVM \cite{7mohr15}&
85.29$\pm$0.49&
81.14$\pm$0.91&
79.69$\pm$0.69&
75.32$\pm$0.41&
78.45$\pm$0.62&
85.46$\pm$0.29\\
Graph-based \cite{3osher15}&
90.82$\pm$1.23&
94.21$\pm$0.83&
95.54$\pm$0.99&
\textbf{95.62$\pm$0.83}&
93.10$\pm$0.78&
86.61$\pm$0.61\\
PCA + Algorithm \ref{alg:Binary} &
83.61$\pm$0.97&
80.12$\pm$0.81&
79.47$\pm$0.91&
82.80$\pm$1.01&
80.52$\pm$0.98&
86.27$\pm$0.88\\
ICA + Algorithm \ref{alg:Binary} &
84.41$\pm$0.93&
82.21$\pm$0.86&
78.88$\pm$0.78&
82.30$\pm$0.99&
83.99$\pm$0.84&
85.57$\pm$1.10\\
APA + SVM &
76.32$\pm$0.78&
77.19$\pm$0.83&
78.61$\pm$0.91&
69.22$\pm$0.87&
73.52$\pm$0.99&
89.90$\pm$0.72\\
Binary APA&
\textbf{98.97$\pm$0.12}&
\textbf{98.17$\pm$0.36}&
\textbf{98.72$\pm$0.16}&
95.26$\pm$0.92&
\textbf{97.23$\pm$0.76}&
\textbf{96.81$\pm$0.79}\\
\hline
\end{tabular}\end{small}\end{center}
\vskip -0.2in
\end{table*}
\begin{table*}[!t]
	\caption{Area under the ROC Curve (AUC) of binary predictors (mean$\pm$std)}
	\label{tbl:AUC}
	\begin{center}\begin{small}\begin{tabular}{lcccccc}
				\hline
				$\downarrow$Alg., Datasets$\rightarrow$ & DS105 (Objects) & DS107 (Words) & DS107 (Consonants)  & DS107 (Objects) & DS107 (Scramble) & DS117 \\
				\hline
SVM \cite{5cox03}&
68.37$\pm$1.01&
67.76$\pm$0.91&
63.84$\pm$1.45&
63.17$\pm$0.59&
66.73$\pm$0.92&
79.36$\pm$0.33\\
Elastic Net \cite{7mohr15}&
78.23$\pm$0.82&
77.94$\pm$0.76&
74.11$\pm$0.82&
81.06$\pm$0.98&
85.54$\pm$0.81&
83.42$\pm$0.68\\
Graph Net \cite{7mohr15}&
77.26$\pm$0.72&
78.31$\pm$0.97&
71.43$\pm$0.58&
82.08$\pm$0.92&
83.97$\pm$0.97&
81.67$\pm$0.74\\
PCA \cite{23otoole05}&
70.69$\pm$0.84&
69.37$\pm$0.77&
65.12$\pm$0.93&
67.56$\pm$0.59&
68.89$\pm$0.90&
79.61$\pm$0.72\\
ICA \cite{39xu13} &
71.33$\pm$0.85&
68.86$\pm$0.93&
71.03$\pm$1.07&
66.91$\pm$0.97&
70.20$\pm$0.72&
78.39$\pm$0.96\\
Selected ROI \cite{6mcmenamin15}&
82.22$\pm$0.42&
86.35$\pm$0.39&
85.63$\pm$0.61&
81.54$\pm$0.92&
85.79$\pm$0.42&
83.71$\pm$0.81\\
L1 Reg. SVM \cite{7mohr15}&
80.91$\pm$0.21&
78.23$\pm$0.57&
77.41$\pm$0.92&
73.92$\pm$0.28&
76.14$\pm$0.47&
83.21$\pm$1.23\\
Graph-based \cite{3osher15}&
88.54$\pm$0.71&
93.61$\pm$0.62&
94.54$\pm$0.31&
94.23$\pm$0.94&
92.23$\pm$0.38&
82.29$\pm$0.91\\
PCA + Algorithm \ref{alg:Binary} &
81.76$\pm$0.90&
78.91$\pm$0.88&
77.44$\pm$0.93&
81.76$\pm$0.12&
77.64$\pm$0.84&
84.32$\pm$0.72\\
ICA + Algorithm \ref{alg:Binary} &
81.11$\pm$0.72&
80.92$\pm$0.58&
75.76$\pm$0.98&
81.04$\pm$0.81&
83.02$\pm$0.92&
82.37$\pm$0.88\\
APA + SVM  &
72.27$\pm$0.86&
73.59$\pm$1.04&
76.95$\pm$0.94&
68.14$\pm$1.02&
71.07$\pm$0.79&
85.10$\pm$0.93\\
Binary APA&
\textbf{97.06$\pm$0.82}&
\textbf{97.31$\pm$0.82}&
\textbf{96.21$\pm$0.62}&
\textbf{94.92$\pm$0.11}&
\textbf{97.21$\pm$0.92}&
\textbf{94.08$\pm$0.84}\\
				\hline
	\end{tabular}\end{small}\end{center}
	\vskip -0.2in
\end{table*}
\begin{table*}[!t]
	\caption{Accuracy of multiclass predictors (mean$\pm$std)}
	\label{tbl:Multiclass}
	\begin{center}\begin{small}\begin{tabular}{lcccc}
				\hline
$\downarrow$Alg., Datasets$\rightarrow$ & DS105 (\# of classes = 8) & DS107 (\# of classes = 4) & ABSTRACT (\# of classes = 5)  & ALL (\# of classes = 10) \\
\hline
Multiclass SVM \cite{5cox03}&
18.03$\pm$4.07&
38.01$\pm$2.56&
31.77$\pm$2.61&
12.26$\pm$5.97\\
MLP \cite{38anderson10}&
38.34$\pm$3.21&
71.55$\pm$2.79&
67.24$\pm$3.72&
32.94$\pm$4.89\\
Selected ROI \cite{6mcmenamin15}&
28.72$\pm$2.37&
68.51$\pm$1.07&
54.19$\pm$2.80&
35.03$\pm$2.66\\
Graph-based \cite{3osher15}&
50.61$\pm$4.83&
89.69$\pm$2.32&
78.96$\pm$3.32&
47.64$\pm$5.28\\
Multiclass APA&
\textbf{59.21$\pm$2.05}&
\textbf{95.61$\pm$1.83}&
\textbf{95.85$\pm$1.05}&
\textbf{62.93$\pm$2.69}\\
\hline
\end{tabular}\end{small}\end{center}
\vskip -0.1in
\end{table*}

The correlation between different ACs can be also meaningful in the feature space. In DS105 and DS107, the scramble (nonsense) stimuli have a low correlation ($>0.13$) in comparison with sensible categories. As another example in DS105, human faces are mostly correlated to the photos of cats and houses (respectively $+0.44$, $+0.25$) in comparison with other objects (the average of correlations is $+0.14$). Another interesting example is the correlation between meaningful stimuli (words and objects) and nonsense stimuli (scrambles and consonants) in DS107, where the meaningful stimuli are highly correlated ($+0.8$) and their correlations with nonsense stimuli are negative (the average of correlation is $-0.65$). Since DS117 is a binary dataset, it is really a good example in order to understand the negative effect of noise and sparsity in fMRI analysis. The correlation between the face category and scramble is around $+0.8$ in the raw voxel space, whereas this correlation is $+0.23$ in the feature space. Indeed, the noisy and sparse raw voxels are not suitable (wise) in order to train a high-performance cognitive model.

Here, we have to note that a suitable feature representation can also improve the performance of the MVPA analysis (the final cognitive model). By considering the geometric analysis of a linear space, classification algorithms draw a hyperplane (that is known as the decision surface in neuroscience \cite{2haxby14}) in a representational space in order to distinguish different classes (categories of visual stimuli). In a highly correlated space, the margin of error for this hyperplane is sensitive. In other words, small changes in the parameters of the hyperplane can rapidly reduce the performance of the classifier in a highly correlated space, such as the raw voxel space \cite{24chen15,25chen16,34tony17b}. Now, suitable feature extraction can minimize the correlation between different categories of visual stimuli. As a result, the margin of error and stability of the final model can be increased in order to train a classifier.

\subsection{Performance Analysis}
In this section, the performance of different methods will be evaluated for both binary and multiclass analyses. In the binary analysis, the performance of the classical binary Support Vector Machine (SVM) is represented. Indeed, this method is used in \cite{5cox03} in order to distinguish different categories of visual stimuli. As regularized methods that are introduced in \cite{7mohr15} for decoding the brain patterns, the performances of L1 regularized SVM, the Elastic Net, and the Graph Net are also reported in this section. Further, the performances of component based methods are also evaluated, i.e. Principal Component Analysis (PCA) that is used in \cite{23otoole05} for training a cognitive model and Independent Component Analysis (ICA), which is employed in \cite{39xu13} in order to analyze fMRI datasets. As another alternative for decoding visual stimuli, the Selected Region of Interest (ROI) method \cite{6mcmenamin15} is reported in this section, where the ROIs for each dataset is manually selected same as the original paper \cite{6mcmenamin15} and then the SVM classifier is applied to the selected ROIs in order to train a cognitive model. As the method was developed in \cite{3osher15}, the performance of a graph-based approach is reported. In order to represent the effect of different parts of the proposed method, we also report three baselines. As the first alternatives, we utilize two component based methods, i.e. `PCA + Algorithm \ref{alg:Binary}' and `ICA + Algorithm \ref{alg:Binary}' that apply Algorithm \ref{alg:Binary} to the features that are respectively extracted by PCA and ICA. As the last baseline, `APA + SVM' applies SVM algorithm to the features that are extracted by APA. In the multiclass analysis, the performance of multiclass SVM is presented as a baseline, where this algorithm was used in \cite{5cox03} in order to generate the cognitive model. Further, the performance of the proposed method is compared with Multilayer Perceptron (MLP) that was introduced as a multiclass approach in \cite{38anderson10} in order to decode the brain patterns. Selected ROI method \cite{6mcmenamin15} and the graph-based approach \cite{3osher15} are also reported as other alternatives. We have to note that a multiclass SVM is used in the Selected ROI method in order to create a multiclass cognitive model (in Table \ref{tbl:Multiclass}). All of the mentioned algorithms are implemented in the MATLAB R2016b ($9.1$) by authors in order to generate experimental results. Further, all evaluations are applied by using leave-one-subject-out cross validation, e.g. we have selected brain patterns of $5$ subjects in DS105 for training a classifier in each iteration and then used the patterns of the rest of the subject in order to test the generated cognitive model. It is worth noting that the number of iterations will be equal to the number of subjects ($5$ in this example). Indeed, not only the brain patterns in training sets and testing sets are independent across subjects but also fMRI data related to each subject was separately preprocessed \cite{29kriegeskorte09}.  We have to note that the same training set and testing set are applied in each iteration to all of the evaluated methods. 

Tables \ref{tbl:BinaryAccuracy} and \ref{tbl:AUC} respectively illustrate the classification Accuracy and Area under the ROC Curve (AUC) for the binary predictors based on the category of the visual stimuli. All visual stimuli in the dataset DS105 except scrambled photos are considered as the object category for generating these experimental results. As these tables depicted, SVM cannot create an acceptable performance on the raw voxels because fMRI data in the original space includes noise and sparsity. Moreover, the performances of the component-based approaches (PCA and ICA) are significantly low because they were applied to whole-brain. Indeed, these methods are suitable for ROI-based problems or rest-mode fMRI data, where they can find the best projection among the wised voxels. Another evidence for this claim is the Selected ROI method, where the performance of this method is significantly improved in comparison with the component-based approaches. In fact, this is the main reason in this paper in order to develop the automatically selected ROI method instead of just applying the component-based methods to the whole-brain data for selecting (ranking) the effective features. As also mentioned in the original paper \cite{7mohr15}, L1 regularized SVM generated better results in comparison with other regularized techniques, i.e. Elastic Net and Graph Net. As another alternative, the graph-based method that is developed by Osher et al. \cite{3osher15} generated acceptable performance because it also employed the anatomical features in order to create a cognitive model. Further, `PCA/ICA + Algorithm \ref{alg:Binary}' have generated better performances in comparison with PCA/ICA methods because of the ensemble approach. Since `APA + SVM' uses better representational space in contrast with the raw fMRI data, it significantly improves the performance of SVM method. Although these baselines can show that each part of the proposed method can generate better performance in comparison with the classical algorithms (PCA, ICA, and SVM), the best results will be produced when we use all parts at the same time. The last but not least, the proposed algorithm has achieved the best performance in comparison with other methods because it provided a better representation of neural activities by exploiting the anatomical structure of the human brain. 

Table \ref{tbl:Multiclass} illustrates the classification accuracy for multiclass predictors. In this table, `DS105' includes $P=8$ different categories (classes) and `DS107' contains $P=4$ categories of the visual stimuli. This paper also combined three datasets in two distinctive forms. `ABSTRACT' includes $5$ different categories, i.e. words, objects, scrambles, consonants, and human faces, which is generated by considering all visual stimuli in the dataset DS105 except faces and scrambled photos as object category and combining them with the datasets DS107 and DS117. Indeed, this combined dataset can be considered for comparing the abstract features of visual stimuli in the human brain. As another alternative, `ALL' in this table generated by combining all of the visual stimuli in the three datasets, i.e. faces, houses, cats, bottles, scissors, shoes, chairs, words, consonants, and scrambled photos. As depicted in Table \ref{tbl:Multiclass}, the accuracy of the proposed method is improved by combining three datasets, whereas the performances of other methods are significantly decreased. As mentioned before, it is the standard space registration problem in the fMRI analysis. In addition, our framework employs the extracted features from the structural regions instead of using all or a subgroup of voxels, which can increase the performance of the predictive models by decreasing noise and sparsity.	

\begin{figure}[t]
	\centering
	\includegraphics[width=0.48\textwidth]{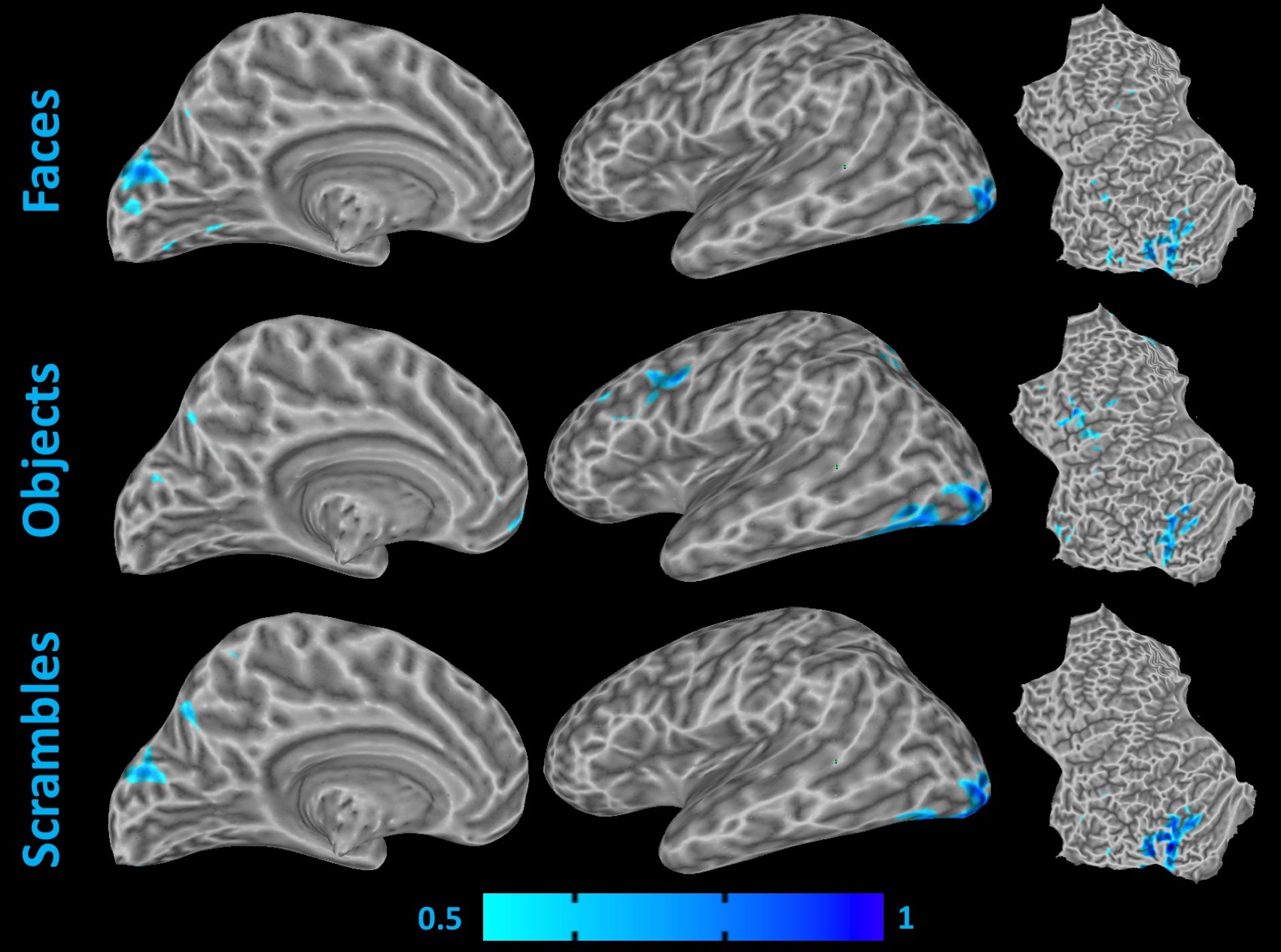}\\	
	\caption{Automatically detected active regions (with Probability greater than 50\%) across abstract categories of visual stimuli, which is generated by combining $3$ datasets, i.e. DS105, DS107, and DS117.}
	\vskip -0.2in
	\label{fig:Regions}
\end{figure}

\section{Discussions and Conclusions}
Anatomical Pattern Analysis (APA) can be used by neuroscientist in order to seek most effective active voxels (regions) across abstract categories of visual stimuli in both a singular dataset and the combined datasets. Figure \ref{fig:Regions} illustrates an example for these active voxels by using the ABSTRACT dataset in Table \ref{tbl:Multiclass} that was generated by combining the visual stimuli in the $3$ datasets, i.e. DS105, DS107, and DS117. In this figure, active regions ($\mathbf{B}^{(\ell)}$) for abstract categories (faces, scrambles, and objects) are normalized in the standard space and then the active voxels with probability greater than 50\% are visualized as the automatically detected ROIs, i.e. Pr$\Big[\sum_{\ell=1}^{U}\tau^{(\ell)}\beta_{.n}^{(\ell)}>50\% \Big]$, where $U$ is the number of all sessions in the combined dataset. As this figure depicted, not only APA can generate a cognitive model in order to predict visual stimuli in the human brain but also it can automatically demonstrate activated loci in human brain across categories of visual stimuli. These activated loci can be used in order to study Specific-Exemplar (SE) recognition \cite{6mcmenamin15} or design an accurate brain mask (ROI) for ROI-based studies \cite{34tony17b}.

In summary, this paper proposes APA framework for decoding visual stimuli in the human brain. This framework uses an anatomical feature extraction method, which provides a normalized representation for combining homogeneous datasets. Further, a new binary imbalance AdaBoost algorithm is introduced. It can increase the performance of prediction by exploiting a supervised random sampling and the correlation between classes. In addition, this algorithm is utilized in an Error-Correcting Output Codes (ECOC) method for multiclass prediction of the brain responses. Empirical studies on $4$ visual categories clearly show the superiority of our proposed method in comparison with the voxel-based approaches. In future, we plan to apply the proposed method to different brain tasks such as low-level visual stimuli, emotion and etc.
\section*{Compliance with Ethical Standards}
\section*{Conflict of Interests }
Muhammad Yousefnezhad and Daoqiang Zhang declare that they have no conflict of interest.
\section*{Ethical Approval}
This article does not contain any studies with human participants or animals performed by any of the authors.
\begin{acknowledgements}
This work was supported in part by the National Natural Science Foundation of China (61422204 and 61473149), and NUAA Fundamental Research Funds (NE2013105).
\end{acknowledgements}


%
%

\bibliographystyle{spmpsci}      
\bibliography{COGN17}   

\begin{thebibliography}{10}
\providecommand{\url}[1]{{#1}}
\providecommand{\urlprefix}{URL }
\expandafter\ifx\csname urlstyle\endcsname\relax
  \providecommand{\doi}[1]{DOI~\discretionary{}{}{}#1}\else
  \providecommand{\doi}{DOI~\discretionary{}{}{}\begingroup
  \urlstyle{rm}\Url}\fi

\bibitem{38anderson10}
Anderson, M., Oates, T.: A critique of multi-voxel pattern analysis.
\newblock In: Proceedings of the Cognitive Science Society, vol.~32 (2010)

\bibitem{22carlson03}
Carlson, T.A., Schrater, P., He, S.: Patterns of activity in the categorical
  representations of objects.
\newblock Journal of Cognitive Neuroscience \textbf{15}(5), 704--717 (2003)

\bibitem{31carroll09}
Carroll, M.K., Cecchi, G.A., Rish, I., Garg, R., Rao, A.R.: Prediction and
  interpretation of distributed neural activity with sparse models.
\newblock NeuroImage \textbf{44}(1), 112--122 (2009)

\bibitem{24chen15}
Chen, P.H., Chen, J., Yeshurun, Y., Hasson, U., Haxby, J., Ramadge, P.J.: A
  reduced-dimension fmri shared response model.
\newblock In: 28th Advances in Neural Information Processing Systems (NIPS-15),
  pp. 460--468. Advances In Neural Information Processing Systems (NIPS),
  December/7--12, MontrÃ©al, Canada (2015)

\bibitem{25chen16}
Chen, P.H., Zhu, X., Zhang, H., Turek, J.S., Chen, J., Willke, T.L., Hasson,
  U., Ramadge, P.J.: A convolutional autoencoder for multi-subject fmri data
  aggregation.
\newblock In: 29th Workshop of Representation Learning in Artificial and
  Biological Neural Networks. Advances In Neural Information Processing Systems
  (NIPS), December/5--10, Barcelona, Spain (2016)

\bibitem{12cohen00}
Cohen, L., Dehaene, S., Naccache, L., Leh{\'e}ricy, S., Dehaene-Lambertz, G.,
  H{\'e}naff, M.A., Michel, F.: The visual word form area: spatial and temporal
  characterization of an initial stage of reading in normal subjects and
  posterior split-brain patients.
\newblock Brain \textbf{123}(2), 291--307 (2000)

\bibitem{19connolly12}
Connolly, A., Gobbini, M., Haxby, J.: Three virtues of similarity-based
  multi-voxel pattern analysis (2012)

\bibitem{20connolly12}
Connolly, A.C., Guntupalli, J.S., Gors, J., Hanke, M., Halchenko, Y.O., Wu,
  Y.C., Abdi, H., Haxby, J.V.: The representation of biological classes in the
  human brain.
\newblock Journal of Neuroscience \textbf{32}(8), 2608--2618 (2012)

\bibitem{5cox03}
Cox, D.D., Savoy, R.L.: Functional magnetic resonance imaging (fmri) `brain
  reading': detecting and classifying distributed patterns of fmri activity in
  human visual cortex.
\newblock NeuroImage \textbf{19}(2), 261--270 (2003)

\bibitem{36duncan09}
Duncan, K.J., Pattamadilok, C., Knierim, I., Devlin, J.T.: Consistency and
  variability in functional localisers.
\newblock NeuroImage \textbf{46}(4), 1018--1026 (2009)

\bibitem{8escalera10}
Escalera, S., Pujol, O., Radeva, P.: Error-correcting output codes library.
\newblock Journal of Machine Learning Research \textbf{11}(Feb), 661--664
  (2010)

\bibitem{4friston03}
Friston, K.J.: Statistical parametric mapping.
\newblock In: Neuroscience Databases, pp. 237--250. Springer (2003)

\bibitem{2haxby14}
Haxby, J.V., Connolly, A.C., Guntupalli, J.S.: Decoding neural representational
  spaces using multivariate pattern analysis.
\newblock Annual Review of Neuroscience \textbf{37}, 435--456 (2014)

\bibitem{14haxby01}
Haxby, J.V., Gobbini, M.I., Furey, M.L., Ishai, A., Schouten, J.L., Pietrini,
  P.: Distributed and overlapping representations of faces and objects in
  ventral temporal cortex.
\newblock Science \textbf{293}(5539), 2425--2430 (2001)

\bibitem{33haxby11}
Haxby, J.V., Guntupalli, J.S., Connolly, A.C., Halchenko, Y.O., Conroy, B.R.,
  Gobbini, M.I., Hanke, M., Ramadge, P.J.: A common, high-dimensional model of
  the representational space in human ventral temporal cortex.
\newblock Neuron \textbf{72}(2), 404--416 (2011)

\bibitem{16haynes06}
Haynes, J.D., Rees, G.: Decoding mental states from brain activity in humans.
\newblock Nature Reviews Neuroscience \textbf{7}(7), 523 (2006)

\bibitem{17haynes07}
Haynes, J.D., Sakai, K., Rees, G., Gilbert, S., Frith, C., Passingham, R.E.:
  Reading hidden intentions in the human brain.
\newblock Current Biology \textbf{17}(4), 323--328 (2007)

\bibitem{35jenkinson02}
Jenkinson, M., Bannister, P., Brady, M., Smith, S.: Improved optimization for
  the robust and accurate linear registration and motion correction of brain
  images.
\newblock Neuroimage \textbf{17}(2), 825--841 (2002)

\bibitem{15kamitani05}
Kamitani, Y., Tong, F.: Decoding the visual and subjective contents of the
  human brain.
\newblock Nature Neuroscience \textbf{8}(5), 679--685 (2005)

\bibitem{11kanwisher97}
Kanwisher, N., McDermott, J., Chun, M.M.: The fusiform face area: a module in
  human extrastriate cortex specialized for face perception.
\newblock Journal of Neuroscience \textbf{17}(11), 4302--4311 (1997)

\bibitem{26kay08}
Kay, K.N., Naselaris, T., Prenger, R.J., Gallant, J.L.: Identifying natural
  images from human brain activity.
\newblock Nature \textbf{452}(7185), 352 (2008)

\bibitem{18kriegeskorte08}
Kriegeskorte, N., Mur, M., Bandettini, P.: Representational similarity
  analysis--connecting the branches of systems neuroscience.
\newblock Frontiers in systems neuroscience \textbf{2} (2008)

\bibitem{29kriegeskorte09}
Kriegeskorte, N., Simmons, W.K., Bellgowan, P.S., Baker, C.I.: Circular
  analysis in systems neuroscience: the dangers of double dipping.
\newblock Nature Neuroscience \textbf{12}(5), 535--540 (2009)

\bibitem{13liesegang02}
Liesegang, T.J.: A cortical area selective for visual processing of the human
  body. downing pe, 1∗ school of psychology, centre for cognitive
  neuroscience, university of wales, bangor, ll57 2as, united kingdom. e-mail:
  p. downing@ bangor. ac. uk jiang y, shuman m, kanwisher n. science 2001; 293:
  2470--2473.
\newblock American Journal of Ophthalmology \textbf{133}(4), 598 (2002)

\bibitem{9liu09}
Liu, X.Y., Wu, J., Zhou, Z.H.: Exploratory undersampling for class-imbalance
  learning.
\newblock IEEE Transactions on Systems, Man, and Cybernetics, Part B
  (Cybernetics) \textbf{39}(2), 539--550 (2009)

\bibitem{10malach95}
Malach, R., Reppas, J., Benson, R., Kwong, K., Jiang, H., Kennedy, W., Ledden,
  P., Brady, T., Rosen, B., Tootell, R.: Object-related activity revealed by
  functional magnetic resonance imaging in human occipital cortex.
\newblock Proceedings of the National Academy of Sciences (PNAS)
  \textbf{92}(18), 8135--8139 (1995)

\bibitem{6mcmenamin15}
McMenamin, B.W., Deason, R.G., Steele, V.R., Koutstaal, W., Marsolek, C.J.:
  Separability of abstract-category and specific-exemplar visual object
  subsystems: Evidence from fmri pattern analysis.
\newblock Brain and Cognition \textbf{93}, 54--63 (2015)

\bibitem{27mitchell08}
Mitchell, T.M., Shinkareva, S.V., Carlson, A., Chang, K.M., Malave, V.L.,
  Mason, R.A., Just, M.A.: Predicting human brain activity associated with the
  meanings of nouns.
\newblock science \textbf{320}(5880), 1191--1195 (2008)

\bibitem{28miyawaki08}
Miyawaki, Y., Uchida, H., Yamashita, O., Sato, M.a., Morito, Y., Tanabe, H.C.,
  Sadato, N., Kamitani, Y.: Visual image reconstruction from human brain
  activity using a combination of multiscale local image decoders.
\newblock Neuron \textbf{60}(5), 915--929 (2008)

\bibitem{7mohr15}
Mohr, H., Wolfensteller, U., Frimmel, S., Ruge, H.: Sparse regularization
  techniques provide novel insights into outcome integration processes.
\newblock NeuroImage \textbf{104}, 163--176 (2015)

\bibitem{1norman06}
Norman, K.A., Polyn, S.M., Detre, G.J., Haxby, J.V.: Beyond mind-reading:
  multi-voxel pattern analysis of fmri data.
\newblock Trends in Cognitive Sciences \textbf{10}(9), 424--430 (2006)

\bibitem{3osher15}
Osher, D.E., Saxe, R.R., Koldewyn, K., Gabrieli, J.D., Kanwisher, N., Saygin,
  Z.M.: Structural connectivity fingerprints predict cortical selectivity for
  multiple visual categories across cortex.
\newblock Cerebral Cortex \textbf{26}(4), 1668--1683 (2015)

\bibitem{23otoole05}
O'toole, A.J., Jiang, F., Abdi, H., Haxby, J.V.: Partially distributed
  representations of objects and faces in ventral temporal cortex.
\newblock Journal of Cognitive Neuroscience \textbf{17}(4), 580--590 (2005)

\bibitem{21rice14}
Rice, G.E., Watson, D.M., Hartley, T., Andrews, T.J.: Low-level image
  properties of visual objects predict patterns of neural response across
  category-selective regions of the ventral visual pathway.
\newblock Journal of Neuroscience \textbf{34}(26), 8837--8844 (2014)

\bibitem{32varoquaux12}
Varoquaux, G., Gramfort, A., Thirion, B.: Small-sample brain mapping: sparse
  recovery on spatially correlated designs with randomization and clustering.
\newblock In: Proceedings of the 29th International Conference on Machine
  Learning (ICML-12), pp. 1375--1382 (2012)

\bibitem{37wakeman15}
Wakeman, D.G., Henson, R.N.: A multi-subject, multi-modal human neuroimaging
  dataset.
\newblock Scientific Data \textbf{2} (2015)

\bibitem{39xu13}
Xu, J., Potenza, M.N., Calhoun, V.D.: Spatial ica reveals functional activity
  hidden from traditional fmri glm-based analyses.
\newblock Frontiers in Neuroscience \textbf{7} (2013)

\bibitem{30yamashita08}
Yamashita, O., Sato, M.a., Yoshioka, T., Tong, F., Kamitani, Y.: Sparse
  estimation automatically selects voxels relevant for the decoding of fmri
  activity patterns.
\newblock NeuroImage \textbf{42}(4), 1414--1429 (2008)

\bibitem{40tony16}
Yousefnezhad, M., Zhang, D.: Decoding visual stimuli in human brain by using
  anatomical pattern analysis on fmri images.
\newblock In: 8th International Conference on Brain Inspired Cognitive Systems
  (BICS'16), pp. 47--57. Springer, November/28--30, Beijing, China (2016)

\bibitem{41tony17a}
Yousefnezhad, M., Zhang, D.: Local discriminant hyperalignment for
  multi-subject fmri data alignment.
\newblock In: 34th AAAI Conference on Artificial Intelligence (AAAI-17), pp.
  59--65. Association for the Advancement of Artificial Intelligence (AAAI),
  February/4--9, San Francisco, California, USA (2017)

\bibitem{34tony17b}
Yousefnezhad, M., Zhang, D.: Multi-region neural representation: A novel model
  for decoding visual stimuli in human brains.
\newblock In: 17th SIAM International Conference on Data Mininig (SDM-17), pp.
  54--62. Society for Industrial and Applied Mathematics (SIAM), April/27--29,
  Houston, Texas, USA (2017)

\end{thebibliography}

%
%

\end{document}